\documentclass[journal]{IEEEtran}

\usepackage[ruled,linesnumbered]{algorithm2e}
\usepackage{amsmath,amssymb,amsfonts}
\usepackage{graphicx}
\usepackage{epstopdf}
\usepackage{subfig}
\usepackage{cite}
\usepackage{gensymb}
\usepackage{subfig}
\usepackage{multirow,booktabs,color,soul,threeparttable}
\usepackage{xcolor}
\usepackage{mdframed}
\usepackage{cleveref}
\newcommand{\RNum}[1]{\uppercase\expandafter{\romannumeral #1\relax}}

\usepackage{amsmath} % For mathematics symbols
\usepackage{array}   % For specifying column formatting
\usepackage{amssymb} % For us
\usepackage{url}
\usepackage{booktabs}
\definecolor{hl}{rgb}{0.75,0.75,0.75}
\sethlcolor{hl}
\definecolor{emph}{rgb}{0,0,1}
\ifCLASSINFOpdf

\else

\fi

\begin{document}

\title{A First Look at Kolmogorov-Arnold Networks in Surrogate-assisted Evolutionary Algorithms}
\author{Hao~Hao, Xiaoqun~Zhang, Bingdong~Li and Aimin~Zhou
% \author{Hao~Hao, Xiaoqun~Zhang, Bingdong~Li and Aimin~Zhou,~\IEEEmembership{Senior~Member,~IEEE}
\thanks{This work was supported in part by the National Natural Science Foundation of China under Grant 62306174; in part by the China Postdoctoral Science Foundation under Grant 2023M742259 and Grant 2023TQ0213; and in part by the Postdoctoral Fellowship Program of the China Postdoctoral Science Foundation under Grant GZC20231588.(Corresponding author: A. Zhou.)}
\thanks{H. Hao, and X. Zhang are with the Institute of Natural Sciences, Shanghai Jiao Tong University, Shanghai 200240, China (e-mail:haohao@sjtu.edu.cn; xqzhang@sjtu.edu.cn).}
\thanks{A. Zhou and B. Li are with the Shanghai Institute of AI for Education, and the School of Computer Science and Technology, East China Normal University, Shanghai 200062, China (e-mail:amzhou@cs.ecnu.edu.cn; bdli@cs.ecnu.edu.cn).}
}

\maketitle

\begin{abstract}
  Surrogate-assisted Evolutionary Algorithm (SAEA) is an essential method for solving expensive expensive problems. Utilizing surrogate models to substitute the optimization function can significantly reduce reliance on the function evaluations during the search process, thereby lowering the optimization costs. The construction of surrogate models is a critical component in SAEAs, with numerous machine learning algorithms playing a pivotal role in the model-building phase. This paper introduces Kolmogorov-Arnold Networks (KANs) as surrogate models within SAEAs, examining their application and effectiveness. We employ KANs for regression and classification tasks, focusing on the selection of promising solutions during the search process, which consequently reduces the number of expensive function evaluations. Experimental results indicate that KANs demonstrate commendable performance within SAEAs, effectively decreasing the number of function calls and enhancing the optimization efficiency. The relevant code is publicly accessible and can be found in the GitHub repository.\footnote{\url{https://github.com/hhyqhh/KAN-EA.git}}
\end{abstract}

\begin{IEEEkeywords}
Kolmogorov-Arnold Networks, expensive optimization, surrogate models, evolutionary algorithms.
\end{IEEEkeywords}

\IEEEpeerreviewmaketitle

\section{Introduction}

In this study, we consider a typical expensive optimization problem defined by a function $f: \mathbb{R}^n \rightarrow \mathbb{R}$, with the aim of finding:
\begin{equation}
\mathbf{x}^* = \arg\min_{\mathbf{x} \in \mathcal{X}} f(\mathbf{x})
\end{equation}
Here, $\mathbf{x} \in \mathbb{R}^n$ represents a vector of decision variables, and $\mathcal{X} \subseteq \mathbb{R}^n$ denotes the feasible region within the decision variable space. The function $f(\mathbf{x})$ returns the value of the objective function for a given vector $\mathbf{x}$, which incurs a high computational cost on evaluation. The optimal solution, $\mathbf{x}^*$, is defined such that $f(\mathbf{x}^*) \leq f(\mathbf{x})$ for all $\mathbf{x} \in \mathcal{X}$.

The challenge in this optimization scenario arises due to the black-box nature of the function $f$, where its explicit form is unknown or undisclosed. This condition renders traditional gradient-based optimization methods ineffective. Each evaluation of the function $f(\mathbf{x})$ typically requires extensive computational resources or significant time, emphasizing the need for optimization methods that minimize the number of function evaluations to reduce overall computational costs. Real-world examples of such expensive optimization issues are extensively witnessed in domains like optimization of prompts for large language models~(LLMs)~\cite{sun2022black}, neural network architecture search~(NAS)~\cite{9508774}, retrosynthetic route planning~\cite{zhang2023evolutionary}, optimization of antenna structures~\cite{hao2022expensive}, and building energy management~\cite{liu2023surrogate}. These applications underscore the relevance and urgent necessity for advanced methods able to efficiently tackle expensive evaluations typical of black-box functions.

Surrogate-assisted evolutionary algorithms (SAEAs) are crucial for solving expensive optimization problems. As gradient-free methods, evolutionary algorithms (EAs) optimize through a trial-and-error approach without relying on the gradient information of the function. Surrogate models approximate and substitute the expensive function effectively, thus reducing the frequency of costly function evaluations during the optimization process and consequently lowering the optimization costs. The development of accurate surrogate models is a key phase in the implementation of SAEAs. The rapid evolution of machine learning (ML) technologies has significantly broadened the spectrum of algorithms available for constructing these models. Prominent algorithms used for surrogate model formation include Gaussian Processes (GP)~\cite{liu2013gaussian}, and Neural Networks (NN)~\cite{pan2018classification}, Radial Basis Functions (RBF)~\cite{yuSurrogateassistedHierarchicalParticle2018} and  Support Vector Machines (SVM)~\cite{HaoZZ21}. Each of these machine learning techniques brings distinct advantages to the table, contributing vastly to the refinement and reliability of surrogate models in evolutionary optimization.

In a recent study~\cite{liu2024kan}, presented shortly before the submission of this paper, a new neural network architecture known as Kolmogorov-Arnold Networks (KANs) was introduced, potentially revolutionizing the role of traditional multilayer perceptrons (MLPs). Inspired by the Kolmogorov-Arnold representation theorem~\cite{kolmogorov1961representation,braun2009constructive}—in contrast to MLPs, which draw on the universal approximation theorem—KANs embody a significant paradigm shift within neural network architectures. This innovative approach differs from common MLP strategies by employing spline-based univariate functions as learnable activation functions instead of conventional linear weights. These adjustments not only refine the accuracy and interpretability of the networks but also allow KANs to achieve comparable or even superior performance on diverse tasks such as data fitting and solving partial differential equations, all while utilizing smaller network structures. Though initially promising in enhancing the efficiency and interpretability of neural network models, the KANs study acknowledges that further empirical research is essential. It is necessary to assess the robustness of KANs across varied datasets and their compatibility with existing deep learning frameworks to fully gauge their potential and limitations.

Following this introduction, subsequent research and applications of KANs have emerged rapidly. Notable developments include integrating KANs into time series analysis~\cite{vacarubio2024kolmogorovarnold}, as well as their incorporation into Recurrent Neural Networks (RNNs)~\cite{genet2024tkan} and a Graph Neural Network variant called GraphKAN~\cite{huagraphkan2024}. These explorations suggest a burgeoning interest and potential for KANs within various subfields of artificial intelligence, indicating a vibrant future trajectory for this novel network architecture in tackling sophisticated computational challenges.

Our paper presents a pioneering study that explores the integration of KANs into surrogate-assisted evolutionary algorithms (SAEAs), an area not previously addressed in the literature. We aim to assess the potential of KANs to enhance the performance of SAEAs, particularly focusing on reducing the number of function evaluations and improving overall optimization efficiency. This integration is tested through the frameworks of Surrogate Pre-selection (SPS) and Surrogate-assisted Selection (SAS), alongside two evolutionary operations: Differential Evolution and Estimation of Distribution.
 
The primary contributions of this work are outlined as follows:
 
\begin{itemize}
  \item Introducing KANs as surrogate models within evolutionary algorithms for the first time, exploring their potential application in solving expensive optimization problems.
  \item Applying KANs to both regression and classification tasks, and integrating these within two evolutionary algorithm frameworks.
  \item Empirically validating the effectiveness of KANs in SAEAs, demonstrating their potential in reducing the number of function evaluations and enhancing optimization efficiency.
\end{itemize}

The structure of our paper proceeds as follows in Section~\ref{sec:preliminaries}, where we introduce the theoretical foundation of KANs and SAEAs. Section~\ref{sec:framework} presents the framework of KANs-assisted evolutionary algorithms, detailing the integration of KANs as surrogate models within EAs. Section~\ref{sec:experiments} provides experimental results and analyses, evaluating the performance of KANs in SAEAs. Finally, Section~\ref{sec:conclusion} concludes the paper with a summary of our findings and future research directions.

\section{Preliminaries}
\label{sec:preliminaries}

\subsection{Kolmogorov-Arnold Networks}

In this section, we introduce the underlying theoretical framework and architecture relevant to our study, namely the Kolmogorov-Arnold Networks (KANs), and their foundation in the Kolmogorov-Arnold theorem.

\subsubsection{Kolmogorov-Arnold Theorem}
KANs diverge significantly from conventional Multi-Layer Perceptrons (MLPs)~\cite{1986Learning}, which rely predominantly on the universal approximation theorem. Instead, KANs are based on the Kolmogorov-Arnold representation theorem~\cite{kolmogorov1961representation}, a key principle in the fields of dynamical systems and ergodic theory. The theorem was independently formulated by Andrey Kolmogorov and Vladimir Arnold in the mid-20th century and asserts a fundamental capability of multivariate functions:
Any continuous multivariate function $f$ dependent $\mathbf{x} = [x_1, x_2, \ldots, x_n] $ within a bounded domain can be represented through a finite composition of simpler univariate functions. This is formally expressed as:
\begin{equation}
    f(\mathbf{x}) = \sum_{i=1}^{2n+1} \Phi_i\left(\sum_{j=1}^n \phi_{i,j}(x_j)\right),
\end{equation}
where \( \Phi_i: \mathbb{R} \to \mathbb{R} \) and \( \phi_{i,j}: [0,1] \to \mathbb{R} \) denote the outer and inner functions, respectively.

\subsubsection{Implementation of Kolmogorov-Arnold Networks}

Recent advancements, particularly the study highlighted in \cite{liu2024kan}, have leveraged this theorem for pioneering a novel neural network architecture. KAN implementation involves a two-layer structure, where each layer deals with simplifications of high-dimensional function approximations into manageable univariate function learning using splines, specifically B-splines. This mathematical model is illustrated in the equation:
\begin{equation}
    f(\mathbf{x}) = \Phi_2(\Phi_1(\phi_{1,1}(x_1), \ldots, \phi_{1,n}(x_n))), 
\end{equation}
suggesting a framework where each function \(\phi\) can be finely tuned through spline interpolation.

In a typical KAN framework, layers are represented by a matrix $\Phi$ consisting of trainable spline functions, seamlessly interconnecting inputs to outputs across the network's architecture. Here, each layer, or KAN block, contributes to the overall learning process by capturing and refining input features through an intricate web of simplified function approximations. This is particularly advantageous in dealing with multivariate function complexities by segregating them into a spectrum of univariate functions which are easier to manage and optimize.

Advanced configurations of KANs, which include widening or deepening the network, enhance its capacity for learning complex features and improve the granularity of function approximation across various datasets. These modifications maintain network differentiability, enabling the use of traditional backpropagation for training and optimization. By merging MLP-like feature extraction with spline-based precision in function approximation, KANs form a robust machine learning architecture. This synthesis effectively combines the high-dimensional processing power of neural networks with the precise, smooth interpolation of splines. Optimizing this advanced architecture could potentially enhance performance in applications that demand high accuracy in function approximation and detailed feature analysis.

\subsection{Surrogate-assisted Evolutionary Algorithms}

Surrogate-assisted evolutionary algorithms (SAEAs) are optimization frameworks that utilize machine learning algorithms to create surrogate models of expensive functions from previously evaluated data. This approach lessens the reliance on direct evaluations of these computationally expensive functions, enhancing optimization efficiency. Numerous machine learning algorithms have been employed to develop these models, significantly improving the optimization process~\cite{hao2020binary}. The integration of surrogate models with evolutionary algorithms is a key area of research in this field. Researchers are focused on effectively melding these models with EAs to enhance performance. The process of select potential solutions is generally categorized into three approaches: regression-based, classification-based, and relation-based methods\cite{hao2022expensive}. 

In regression-based approaches, such as the ones proposed by Jones et al.~\cite{jones1998efficient}, Liu et al.~\cite{liu2014gaussian}, and Li et al.~\cite{li2020surrogate}, surrogate models like Gaussian processes and radial basis functions predict continuous outputs and are used to enhance optimization efficiency and convergence within genetic and differential evolution algorithms. Classification-based models, discussed by Zhou et al.~\cite{zhou2019fuzzy} and Wei et al.~\cite{wei2020classifier}, focus on mapping input vectors to discrete classes to discern solution quality, improving algorithms' performance through efficient filtering mechanisms. Lastly, relation-based surrogates, as utilized by Hao et al.~\cite{hao2020binary, hao2024enhancing} and Yuan et al.~\cite{yuan2021expensive}, learn and predict relationships among solutions to directly enhance the select process in both single and multi-objective optimization problems, demonstrating significant advancements in algorithm performance and efficiency. 

While a substantial body of work has focused on integrating surrogate models with EAs to enhance optimization processes, recent years have seen a scarcity of novel machine learning techniques being introduced into this field. However, KANs, as a new class of neural network architecture, hold significant potential for transformative impacts. This paper introduces KANs into surrogate-assisted evolutionary algorithms (SAEAs), exploring their application and efficacy in addressing expensive optimization problems.

\section{KANs-assisted EA Framework}
\label{sec:framework}

In this section, we introduce the framework of KANs-assisted evolutionary algorithms, providing a guide on the integration and utilization of KANs as surrogate models within EAs. We begin with an overview of the overall framework, followed by a detailed discussion of the specific steps involved in implementing KANs as surrogate models. Lastly, we present two concrete examples of KANs-assisted EAs, specifically focusing on Surrogate Pre-selection (SPS) and Surrogate-assisted Selection (SAS) Framework.

\begin{figure}[htbp!]
  \centering
  \includegraphics[width=1\linewidth]{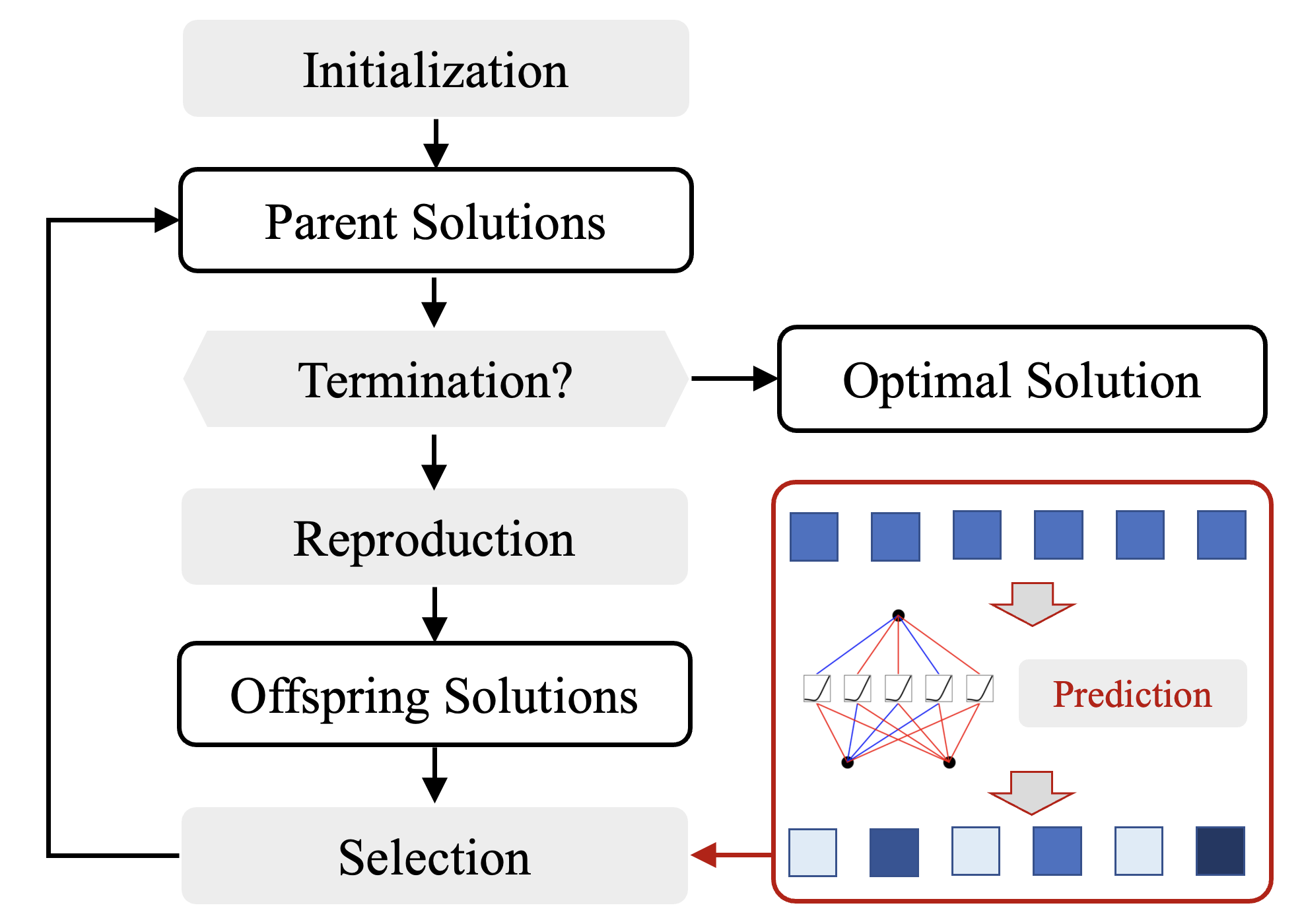}
  \caption{Framework of KANs-assisted EA.}
  \label{fig:framework}
\end{figure}

\subsection{Framework of KANs-assisted EA}

Fig.~\ref{fig:framework} illustrates an overview of the framework for an EA-assisted by KANs. This structure includes several key phases:
\begin{itemize}
  \item Initialization: The process begins with the initialization of a population, where a population generated through random sample operators.
  \item Reproduction: Parent individuals from the existing population produce offspring using reproduction operators, such as crossover and mutation. These genetic operators are designed to explore the solution space by combining and altering the genetic material of parent solutions.
  \item Model-assisted Selection: Instead of evaluating all offspring using the expensive function, a surrogate model predicts the quality of the new solutions. In the context of KANs-assisted EA, KANs serve as the surrogate model. This model approximates the objective function or label based on previously evaluated data and predicts the performance of offspring. Then, a subset of offspring is selected and evaluated by the expensive function.
  % \item Re-evaluation: The surrogate model assists in selecting the most promising offspring to form the next generation. This model-based selection process critically reduces the number of expensive function evaluations by filtering out less promising candidates before they are actually evaluated.
  \item Termination: The algorithm iterates through these steps until predefined termination criteria are met, such as a maximum number of generations or a sufficient solution quality.
  \item Output: The algorithm outputs the optimal solution, which is the best solution found across all generations according to the surrogate model’s predictions and real evaluations performed.
\end{itemize}

The role of surrogate-assisted selection within this framework is crucial, it predicts the quality of solutions in the absence of true evaluations. This predictive capacity enables the algorithm to make informed decisions about which solutions to carry forward into future generations, significantly economizing on computational resources by reducing the dependency on costly objective function evaluations.

\subsection{KANs as Surrogates}

In Surrogate-assisted Evolutionary Algorithms (SAEAs), the surrogate model involves two main processes: training and prediction. During the training phase, the surrogate model is constructed using a set of solutions $\mathcal{D}$ that have already been evaluated. This dataset $\mathcal{D}$ provides the necessary input-output pairs (solution vectors and their corresponding function values or categories) that the surrogate model uses to learn the underlying patterns of the optimization problem. In the prediction phase, the trained surrogate model is then used to predict the quality (either the function values or classes) of a newly generated set of solution candidates $\mathcal{U}$. These predictions help to estimate the potential of these candidates without the need for costly function evaluations.

In the case of Kolmogorov-Arnold Networks (KANs) within this framework, the details are as follows:

\subsubsection{Regression task}
\begin{figure}[htbp!]
  \centering
  \includegraphics[width=.8\linewidth]{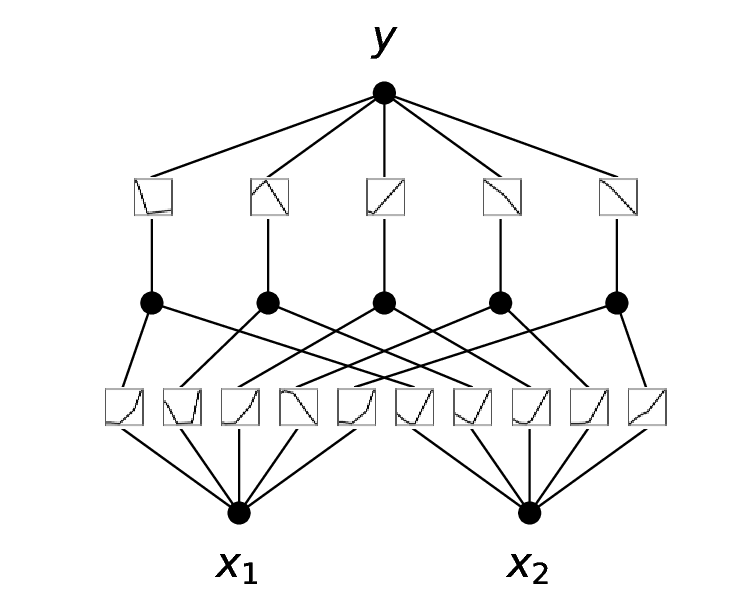}
  \caption{Use KANs as a regression model}
  \label{fig:kan_reg}
\end{figure}

In regression tasks within SAEAs using KANs, the network's architecture is designed to handle the training dataset $\mathcal{D} = \{\mathbf{x}_i, y_i\}$ effectively, where $\mathbf{x}_i$ are the input features and $y_i$ are the target values. Additionally, within the context of EAs, the input features $\mathbf{x}$ are the solution vectors, and the target values $y$ are the objective function values. The KAN setup includes an input layer matching the dimensionality $n$ of features $\mathbf{x}$, hidden layers sized $2n+1$, and a single output layer for continuous target predictions,as shown in Fig.~\ref{fig:kan_reg}. During the training phase, KANs learn dataset patterns by optimizing the mean squared error between predictions and actual values using the L-BFGS method~\cite{liu2024kan}. In the prediction phase, KANs assess a newly generated set of solutions $\mathcal{U}$, predicting each solution's target value.

\subsubsection{Classification task}

\begin{figure}[htbp!]
  \centering
  \includegraphics[width=.8\linewidth]{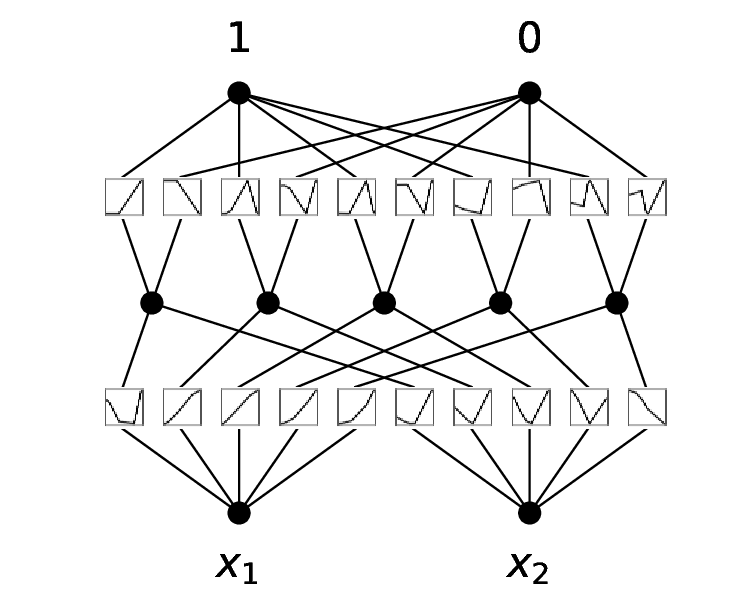}
  \caption{Use KANs as a classification model}
  \label{fig:kan_cla}
\end{figure}

In classification tasks, the architectural framework of KANs is similar to that used in regression, but with a key modification in the output layer where a softmax activation function is employed for categorical output. The structure of KANs includes an input layer matched to the dimension $n$ of the features $\mathbf{x}$, a hidden layer of size $2n+1$, and a single output layer designed for categorical outputs, as illustrated in Figure~\ref{fig:kan_cla}. Specifically, KANs process the training dataset $\mathcal{D} = \{\mathbf{x}_i, l_i\}$, where $\mathbf{x}_i$ are the input features and $l_i$ are the category labels. And the input features $\mathbf{x}$ represent the solution vectors, and the category labels $l$ reflect the quality of the solutions, distinguishing ``good'' from ``bad'' solutions. During the training phase, KANs learn the patterns of the dataset through the minimization of the cross-entropy loss function. In the prediction phase, the trained KANs predict category labels for a newly generated set of solutions $\mathcal{U}$.

\subsection{Implementation of KANs-assisted EA}

In this section, two specific frameworks for integrating Kolmogorov-Arnold Networks (KANs) into surrogate-assisted evolutionary algorithms are presented: the Surrogate Pre-selection (SPS) and Surrogate-assisted Selection (SAS) Frameworks, as show in Table~\ref{tab:alg_imp}. The former focuses on investigating the capability of KANs to predict promising solutions during the pre-selection phase, while the latter examines the effectiveness of KAN-assisted evolutionary algorithms and compares their performance with mainstream algorithms.

\begin{table}[htbp]
\centering
\caption{Two implementations of KANs-assisted EA frameworks.}
\begin{tabular}{@{}lp{3.1cm}p{3.1cm}@{}}
\toprule
\textbf{Algorithm} & \textbf{KAN-SPS} & \textbf{KAN-SAS} \\
\midrule
\textbf{Highlight} & SPS framework requires no extra model management strategies and exhibits the performance benefits of the KANs model. & SAS framework adopts a model management strategy, enhancing KANs performance. \\
% \textbf{Reproduction} & CoDE~\cite{wang2011differential}, generates multiple experimental solutions naturally. & EDA~\cite{journals/tec/ZhouSZ15}, noted for efficient global search. \\
\textbf{Purpose} & To study convergence performance gains by KANs in an EA framework. & To explore specific impacts of KANs-assisted EA. \\
\bottomrule
\end{tabular}
\label{tab:alg_imp}
\end{table}

\subsubsection{Surrogate Pre-selection (SPS) Framework}

\begin{algorithm}[htbp]
  % \caption{KANs Pre-selection (KAN-SPS) algorithm}\label{alg:sps}
  \caption{KANs pre-selection algorithm}\label{alg:sps}
  \SetKwInOut{Input}{Input}\SetKwInOut{Output}{Output}
  \Input{$N$~(population size), $t$~(trial scaling factor), $fes_{max}$~(maximum number of fitness evaluations) }
  \Output{$\mathbf{x}^{*}$~(optimal solution)}  
  \BlankLine
  Initialize $\mathcal{P}=\{\mathbf{x}_1,\ldots,\mathbf{x}_N\}$ and evaluate them by expensive function $f$\; \label{alg:sps:init}
  $fes \leftarrow N$\; \label{alg:sps:fes}
  \While{$fes <fes_{max}$}{  \label{alg:sps:while}
    Train KANs using $\mathcal{P}$\;  \label{alg:sps:train}
    \For{each $\mathbf{x} \in \mathcal{P}$}{  
      Generate $t$ trial solutions $\mathbf{u}_1,\ldots,\mathbf{u}_t$ using CoDE operators\;    \label{alg:sps:gen}
      Estimate $\{\mathbf{u}_1,\ldots,\mathbf{u}_t \}$ by KANs\ and select the best one $\mathbf{u}^*$\;  \label{alg:sps:preselect}
      Evaluate $\mathbf{u}^*$ by expensive function $f$\; \label{alg:sps:eval}
      \If{$f(\mathbf{u}^*) < f(\mathbf{x})$}{   
        $\mathbf{x} \leftarrow \mathbf{u}^*$\;  \label{alg:sps:replace}
      }
    $fes \leftarrow fes + 1$\;  \label{alg:sps:upfes}
    }
  }
\end{algorithm}

Algorithm~\ref{alg:sps} illustrates the process of pre-selection using Kolmogorov-Arnold Networks (KANs), and the specific details are described as follows:
\begin{itemize}
  \item \textbf{Initialization} (lines~\ref{alg:sps:init}-\ref{alg:sps:fes}): At the beginning of the algorithm, a population $\mathcal{P}$ is initialized and evaluated using the expensive function $f$.
  \item \textbf{Training KANs} (line~\ref{alg:sps:train}): At the start of each generation, KANs are trained using the current population $\mathcal{P}$. This training could be for either a classification or a regression task. In classification, solutions are categorized based on their objective values, with the top 50\% labeled as class ``1'' and the bottom 50\% as class ``0''. In regression, KANs learn to predict the actual objective values.
  \item \textbf{Generating Trial Solutions} (line~\ref{alg:sps:gen}): For each solution $\mathbf{x} \in \mathcal{P}$, $t$ trial solutions $\mathbf{u}_1,\ldots,\mathbf{u}_t$ are generated using the CoDE operator~\cite{wang2011differential}. The CoDE operator is a composite differential evolution strategy that enhances search performance by integrating multiple strategies.
  \item \textbf{Estimating Trial Solutions} (line~\ref{alg:sps:eval}): Candidate solutions are evaluated using predictions from KANs, and the potential optimal solution $u^*$ is selected based on KAN's predictions. For regression tasks, the solution with the minimum predicted value is chosen, whereas for classification tasks, solutions labeled as 1 are preferred. If there is more than one, one is randomly selected.
  \item \textbf{Replacement} (line~\ref{alg:sps:replace}): The selected $\mathbf{u}^*$ is assessed using the expensive function $f$. If the objective value of $u^*$ is better than that of the current solution $\mathbf{x}$, then the current solution is replaced by $\mathbf{u}^*$.
\end{itemize}

This framework is characterized by generating more candidate solutions without additional evaluation overhead; additionally, it operates without extra model management strategies, thus clearly demonstrating the performance benefits introduced by incorporating the KANs model. For clarity in discussion, this algorithm is referred to as the KANs Pre-selection (KAN-SPS) algorithm. Two versions of the algorithm are delineated based on their operational tasks: regression and classification. These variants are denoted as KAN-SPS-reg for the regression version and KAN-SPS-cla for the classification version.

\subsubsection{Surrogate-assisted Selection (SAS) Framework}

\begin{algorithm}[htbp]
  \caption{KANs-assisted selection algorithm}\label{alg:sas}
  \SetKwInOut{Input}{Input}\SetKwInOut{Output}{Output}
  \Input{$N$ (population size), $\tau$ (training data size), $fes_{max}$ (maximum number of fitness evaluations)}
  \Output{$\mathbf{x}^{*}$ (optimal solution)}
  \BlankLine
  Initialize $\mathcal{P}_e = \{\mathbf{x}_1, \ldots, \mathbf{x}_N\}$ and evaluate them using the expensive function $f$\; \label{alg:sas:init}
  $\mathcal{P}_{u} \leftarrow \emptyset$\;
  $\mathcal{A} \leftarrow \mathcal{P}_e$\; \label{alg:sas:save}
  $fes \leftarrow N$\; \label{alg:sas:fes}
  \While{$fes < fes_{max}$}{ \label{alg:sas:while}
    Train KANs using $\mathcal{A}_{1:\tau}$\; \label{alg:sas:train}
    Generate offspring $\mathcal{O}$ using EDA operators from $\mathcal{P}_e\cup \mathcal{P}_u$\; \label{alg:sas:gen}
    Use KANs to select the best solution $o^{*}$ and the top $N/2$ high-quality solutions $\mathcal{P}_{u}$ from $\mathcal{O}$\; \label{alg:sas:select}
    Evaluate $o^{*}$ using the expensive function $f$, and merge $o_{*}$ to $\mathcal{A}$\; \label{alg:sas:eval}
    Select the top $N$ solutions from $\mathcal{A}$ as the next generation population $\mathcal{P}_e$\; \label{alg:sas:replace}
    $fes \leftarrow fes + 1$\; \label{alg:sas:upfes}
  }
\end{algorithm}

In the above implementation, due to the lack of an effective model management strategy, the advantages of KANs were not fully realized. Therefore, this paper will implement a KANs-assisted evolutionary algorithm with a model management strategy, as shown in Algorithm~\ref{alg:sas}. The specific details are described as follows:

\begin{itemize}
  \item \textbf{Initialization} (lines~\ref{alg:sas:init}-\ref{alg:sas:fes}): A population $\mathcal{P}_e$ is initialized and evaluated by the expensive function $f$. The evaluated population is saved in $\mathcal{A}$, and an empty set $\mathcal{P}_u$ is prepared to store unevaluated solutions, updating the number of function evaluations $fes$ to $N$.
  
  \item \textbf{Model Training} (line~\ref{alg:sas:train}): The KANs models are trained using the top $\tau$ solutions from $\mathcal{A}$. It can involve training a regression model to predict objective values or a classification model where the top 30\% of solutions are labeled as '1' and the bottom 70\% as '0'.
  
  \item \textbf{Generating offsprings} (line~\ref{alg:sas:gen}): EDA (Estimation of Distribution Algorithm) operators are utilized to generate new offsprings $\mathcal{O}$, employing the Variable Width Histogram (VWH)~\cite{} from both evaluated population $\mathcal{P}_e$ and unevaluated pool $\mathcal{P}_u$ due to their efficient convergence properties~\cite{hao2024enhancing}.
  
  \item \textbf{Model-Assisted Selection} (line~\ref{alg:sas:select}): KANs then aid in choosing the best possible solution $\mathcal{o}^*$ from the offspring $\mathcal{O}$, along with the top $N/2$ high-quality candidates for $\mathcal{P}_u$. The selection criterion depends on the model type; regression tasks might select the solutions with the lowest predicted values, while classification tasks selected solutions labeled as '1'.

  \item \textbf{Re-evaluation and Update} (lines~\ref{alg:sas:eval}-\ref{alg:sas:replace}): The solution $o^*$ is reevaluated using the expensive function $f$. It is then added to $\mathcal{A}$. The top $N$ solutions from $\mathcal{A}$ are selected to form the next generation population $\mathcal{P}_e$.

\end{itemize}
The model management strategy mentioned above has already been validated for efficiency~\cite{hao2024model}. The algorithm is referred to as the KAN-assisted Selection Algorithm (KAN-SAS). The version that uses only regression models is called KAN-SAS-I, and the version that uses both regression and classification models to select $\mathbf{o}^*$ and $\mathcal{P}_{u}$ is called KAN-SAS-II.

\section{Empirical Studies}
\label{sec:experiments}

This section will analyze the performance of KANs in surrogate-assisted evolutionary algorithms. First, it demonstrates the advantages of KANs over MLP in regression and classification tasks through 2D visualization, especially under small-scale network structures. Then, within the frameworks of pre-selection and selection, it evaluates the performance of KANs on four classic optimization problems.

\subsection{Test suites}
In this work, four typical single-objective functions are selected for testing: the Ellipsoid function, Rosenbrock function, Ackley function, and Griewank function. These functions are widely utilized in the field of optimization~\cite{liu2013gaussian} and exhibit varied characteristics such as convexity, non-convexity, and multimodality, among others. The specific definitions of these functions are presented in Table~\ref{tab:benchmark_functions}.

\begin{table*}[htbp]
  \centering
  \caption{Benchmark functions in the empirical study.}
  \renewcommand{\arraystretch}{1.5}
  \begin{tabular}{|>{\raggedright\arraybackslash}m{1.3cm}|>{\centering\arraybackslash}m{6cm}|>{\centering\arraybackslash}m{3cm}|>{\centering\arraybackslash}m{5cm}|m{3cm}|}
  \hline
  \textbf{Name} & \textbf{Formula}  & \textbf{Boundaries}  & \textbf{Description } \\
  \hline
  Ellipsoid & 
  $f(\mathbf{x}) = \sum_{i=1}^{n} i \cdot x_i^2$
   & $x \in[-5.12, 5.12]^n$ & A convex quadratic function with its global minimum at the origin.\\
  \hline
  Rosenbrock & 
  $f(\mathbf{x}) = \sum_{i=1}^{n-1} \left[100 (x_{i+1} - x_i^2)^2 + (x_i - 1)^2\right]$ 
 & $x \in[-2.048, 2.048]^n$ & A non-convex function with a challenging narrow, curved valley. \\
  \hline
  Ackley & 
  $f(\mathbf{x}) = -20 \exp\left(-0.2 \sqrt{\frac{1}{n} \sum_{i=1}^{n} x_i^2}\right) - \exp\left(\frac{1}{n} \sum_{i=1}^{n} \cos(2\pi x_i)\right) + 20 + e$ 
  & $x \in[-32.768, 32.768]^n$ & Features a flat outer region and a large hole at the center with many local minima. \\
  \hline
  Griewank & 
  $f(\mathbf{x}) = 1 + \frac{1}{4000} \sum_{i=1}^{n} x_i^2 - \prod_{i=1}^{n} \cos\left(\frac{x_i}{\sqrt{i}}\right)$
  & $x \in[-600, 600]^n$ & Contains many local minima with oscillatory behaviour through cosine terms. \\
  \hline
  \end{tabular}
  \label{tab:benchmark_functions}
\end{table*}

\subsection{2D Visualization}

This section compares KANs with traditional Multi-Layer Perceptrons (MLPs) by visualizing their performance on regression and classification tasks using 2-dimensional functions. The chosen functions for this demonstration are listed in Table~\ref{tab:benchmark_functions}, with a dimensionality set to two. A set of 50 random samples is used as training data to train both KANs and MLP models. Predictions are then made across the entire area and visualized to showcase the effectiveness of the models. Both KANs and MLPs employ an identical network architecture featuring a hidden layer whose width is 2n+1, where n represents the input dimension. The models are iterated 50 times for training.

\begin{figure*}[htbp!]
\centering
\includegraphics[width=.8\linewidth]{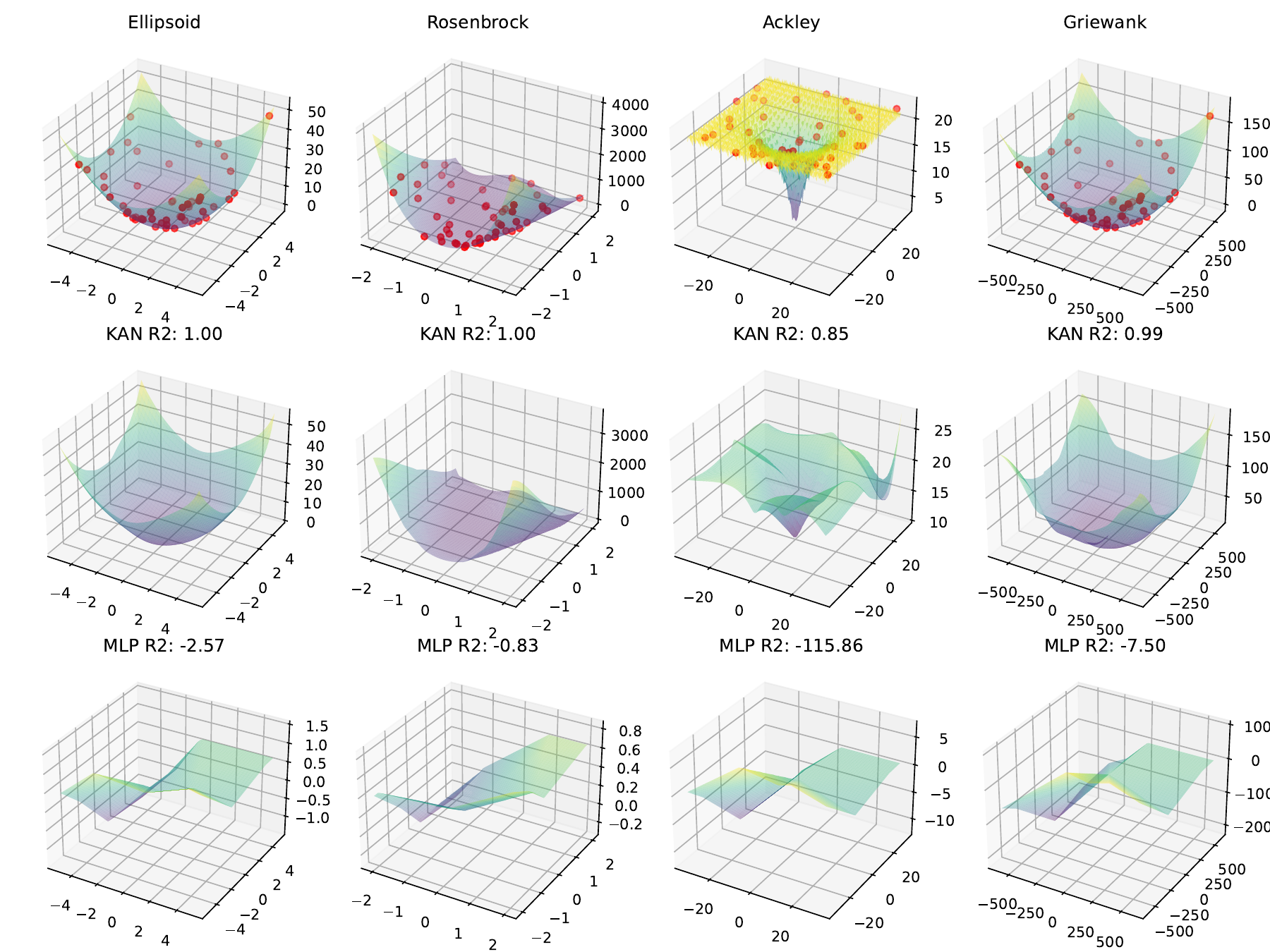}
\caption{Performance of KANs and MLPs on 2D regression tasks. In the first row, the red dots represent the training data, and the surface illustrates the true function. In the second and third rows, the surfaces represent the predicted results from KANs and MLPs, respectively. The R2 score is used to measure the accuracy of the fit, with values closer to 1 indicating better fitting performance.}
\label{fig:2d_reg}
\end{figure*}

Fig.~\ref{fig:2d_reg} displays the regression performance, evaluated using the R2 score. The results indicate that under conditions of limited training data and simple network structures, KANs consistently outperform MLPs across all four functions. The MLPs, on the other hand, show near failure in these tasks.

\begin{figure*}[htbp!]
\centering
\includegraphics[width=.8\linewidth]{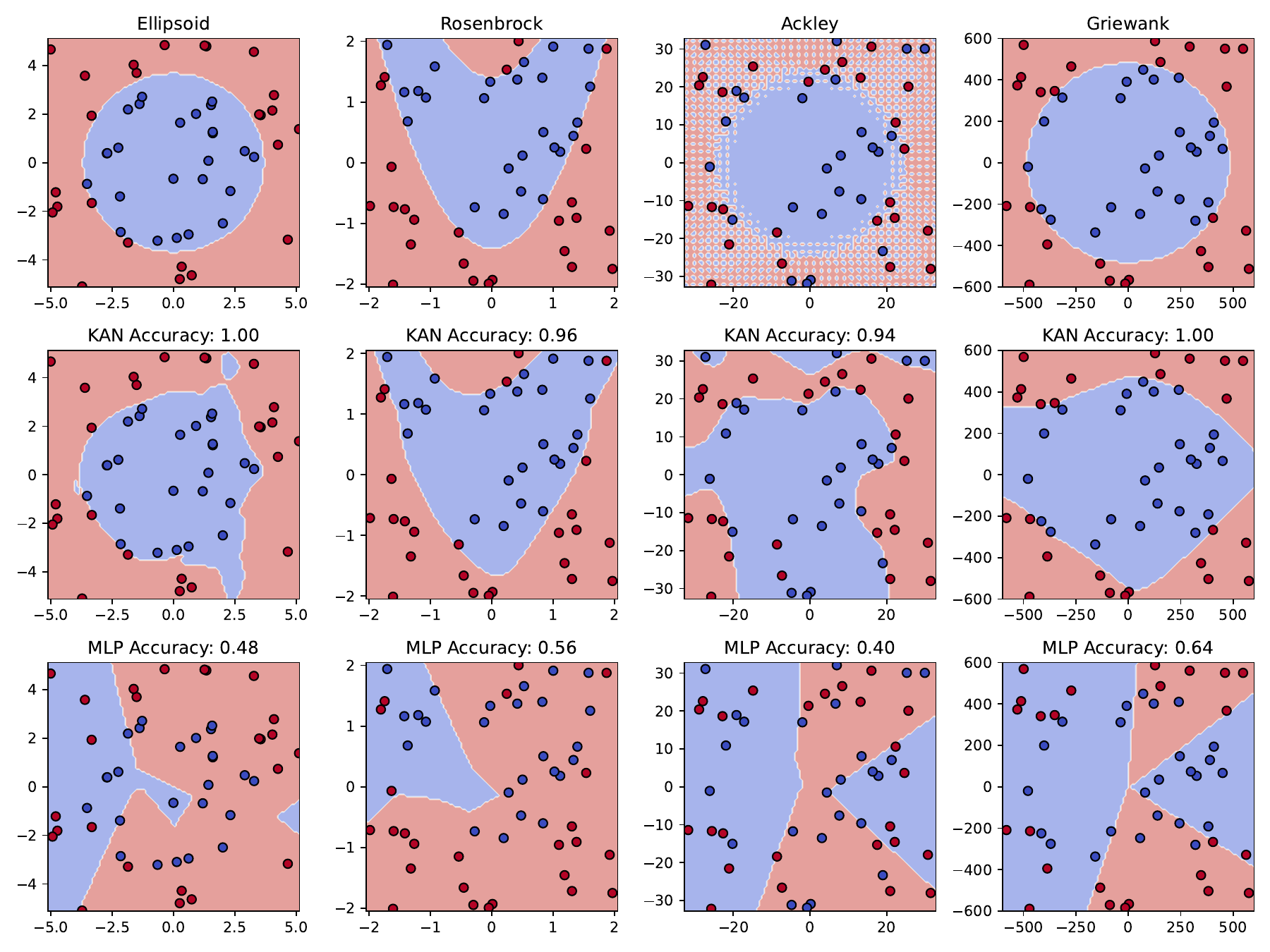}
\caption{Performance of KANs and MLPs on 2D classification tasks. In the figure, the dots represent the training data, with red and blue colors used to distinguish the categories. In the first row, the differently colored shaded areas indicate the true distribution of classes. In the second and third rows, the shaded areas represent the predictive results of KANs and MLPs, respectively.}
\label{fig:2d_cla}
\end{figure*}

Fig.~\ref{fig:2d_cla} illustrates the performance on classification tasks, where the sampled 50 data points are split into two categories based on their function values. Both KANs and MLP models are trained with this data, followed by predictions across the entire region, which are then visualized to demonstrate their predictive efficacy. The results indicate that KANs significantly outperform MLPs in classification tasks as well.

\subsection{Study on KAN-SPS}

This section investigates the performance of KANs within the SPS framework. For comparison, an MLP with the same three-layer structure is used as a control group, and a variant randomly selected from the trial solutions serves as the baseline. The specific configurations are as follows:
 
\begin{itemize}
 \item KAN-SPS-reg: Uses KANs as the regression model for pre-selection.
 \item KAN-SPS-cla: Uses KANs as the classification model for pre-selection.
 \item MLP-SPS-reg: Uses MLP as the regression model for pre-selection.
 \item MLP-SPS-cla: Uses MLP as the classification model for pre-selection.
 \item SPS-random: Randomly selects a solution from the trial solutions (Algorithm~\ref{alg:sps} line \ref{alg:sps:preselect}).
\end{itemize}
 
Experimental parameters are designed as follows:
\begin{itemize}
 \item Population size: $N=50$.
 \item Maximum number of evaluations: $fes_{max}=2000$.
 \item Number of trial solutions per generation: $t=3$.
 \item Problem dimensions: $n=5, 10$.
\end{itemize}
 
Each experiment is repeated 30 times to assess the stability of the algorithm. Table~\ref{tab:kan-sps} displays the comparisons of minimum values between KAN-SPS and other algorithms across 30 independent runs. The best results are highlighted in bold. The Wilcoxon rank-sum test~\cite{hollander2013nonparametric} is used to statistically analyze the results and evaluate significant differences between the algorithms. The results indicate that both KAN-SPS-reg and KAN-SPS-cla outperform the other algorithms in all tested instances, achieving average rank values of 1 and 2.375, respectively. The Wilcoxon rank-sum test shows that KAN-SPS-reg is significantly superior to the other algorithms across all eight test problems.

\begin{table*}[htbp]
\renewcommand{\arraystretch}{1.1}
\renewcommand{\tabcolsep}{4pt}
\centering
\caption{Comparison of minimize values across 30 independent runs between KAN-SPS and other algorithms. The best results are highlighted in bold.}
\scriptsize
\begin{tabular}{ccccccc}
\toprule
problem &$n$ & KAN-SPS-reg & KAN-SPS-cla & MLP-SPS-reg & MLP-SPS-cla & SPS-random \\
\midrule
\multirow{4}{*}{Ellipsoid} &\multirow{2}{*}{5}& \hl{3.95e-04[1]} & 1.81e-02[2]($-$) & 1.01e-01[3]($-$) & 1.49e-01[4]($-$) & 2.12e-01[5]($-$) \\ 
&& (1.45e-04) & (8.08e-03) & (4.10e-02) & (9.93e-02) & (1.48e-01) \\  \cline{2-7}
&\multirow{2}{*}{10}& \hl{1.97e+00[1]} & 5.90e+00[2]($-$) & 1.33e+01[3]($-$) & 1.53e+01[4]($-$) & 1.71e+01[5]($-$) \\ 
&& (8.50e-01) & (1.62e+00) & (3.50e+00) & (5.04e+00) & (5.79e+00) \\  \hline
\multirow{4}{*}{Rosenbrock} &\multirow{2}{*}{5}& \hl{1.13e+00[1]} & 1.70e+00[2]($-$) & 3.93e+00[3]($-$) & 4.33e+00[4]($-$) & 4.76e+00[5]($-$) \\ 
&& (5.24e-01) & (5.20e-01) & (9.70e-01) & (1.33e+00) & (1.29e+00) \\  \cline{2-7}
&\multirow{2}{*}{10}& \hl{2.37e+01[1]} & 5.40e+01[2]($-$) & 8.87e+01[3]($-$) & 8.91e+01[4]($-$) & 9.39e+01[5]($-$) \\ 
&& (7.52e+00) & (7.82e+00) & (3.64e+01) & (2.37e+01) & (3.17e+01) \\  \hline
\multirow{4}{*}{Ackley} &\multirow{2}{*}{5}& \hl{7.70e-01[1]} & 1.69e+00[2]($-$) & 3.41e+00[3]($-$) & 4.78e+00[5]($-$) & 4.33e+00[4]($-$) \\ 
&& (2.23e-01) & (3.27e-01) & (6.26e-01) & (1.06e+00) & (7.59e-01) \\  \cline{2-7}
&\multirow{2}{*}{10}& \hl{7.26e+00[1]} & 1.11e+01[2]($-$) & 1.18e+01[3]($-$) & 1.33e+01[5]($-$) & 1.31e+01[4]($-$) \\ 
&& (1.07e+00) & (9.62e-01) & (1.26e+00) & (1.48e+00) & (6.93e-01) \\  \hline
\multirow{4}{*}{Griewank} &\multirow{2}{*}{5}& \hl{4.39e-01[1]} & 6.50e-01[2]($-$) & 1.11e+00[3]($-$) & 1.14e+00[4]($-$) & 1.25e+00[5]($-$) \\ 
&& (1.12e-01) & (9.15e-02) & (2.04e-01) & (2.53e-01) & (2.18e-01) \\  \cline{2-7}
&\multirow{2}{*}{10}& \hl{6.74e+00[1]} & 1.64e+01[5]($-$) & 1.23e+01[2]($-$) & 1.52e+01[3]($-$) & 1.60e+01[4]($-$) \\ 
&& (2.26e+00) & (5.54e+00) & (2.53e+00) & (5.10e+00) & (3.73e+00) \\  \hline
mean rank && 1.00 & 2.375 & 2.875 & 4.125 & 4.626 \\ 
$+$ / $-$ / $\approx$ &&  & 0/8/0 & 0/8/0 & 0/8/0 & 0/8/0 \\ 
\bottomrule
\end{tabular}
\label{tab:kan-sps}
\end{table*}

\begin{table*}[htbp]
  \renewcommand{\arraystretch}{1.1}
  \renewcommand{\tabcolsep}{4pt}
  \centering
  \caption{Comparison of minimize values across 30 independent runs between KAN-SAS and other algorithms. The best results are highlighted in bold.} \scriptsize
  \begin{tabular}{ccccccc}
  \toprule
  problem & $n$&KAN-SAS-I & KAN-SAS-II & XGB-SAS & RF-SAS & BO \\
  \midrule
  \multirow{4}{*}{Ellipsoid} &\multirow{2}{*}{5}& \hl{1.08e-06[1]} & 7.07e-05[2]($-$) & 1.17e-03[3]($-$) & 1.15e-02[5]($-$) & 6.15e-03[4]($-$) \\ 
  & & (7.90e-07) & (4.96e-05) & (2.93e-03) & (1.94e-02) & (2.65e-03) \\  \cline{2-7}
  &\multirow{2}{*}{10}& \hl{3.73e-02[1]} & 5.94e-02[2]($\approx$) & 3.89e-01[3]($-$) & 1.06e+00[4]($-$) & 1.51e+00[5]($-$) \\ 
  && (1.93e-02) & (3.95e-02) & (1.65e-01) & (8.21e-01) & (3.24e+00) \\  \hline
  \multirow{4}{*}{Rosenbrock} &\multirow{2}{*}{5}& 3.54e+00[3] & \hl{2.57e+00[1]($+$)} & 2.78e+00[2]($\approx$) & 4.73e+00[4]($\approx$) & 1.08e+01[5]($-$) \\ 
  && (1.55e+00) & (3.43e-01) & (1.08e+00) & (3.43e+00) & (6.82e+00) \\   \cline{2-7}
  &\multirow{2}{*}{10}& \hl{1.06e+01[1]} & 1.89e+01[2]($\approx$) & 2.97e+01[3]($-$) & 3.50e+01[4]($-$) & 9.45e+01[5]($-$) \\ 
  && (1.61e+00) & (1.31e+01) & (1.78e+01) & (1.49e+01) & (2.20e+01) \\  \hline
  \multirow{4}{*}{Ackley} &\multirow{2}{*}{5}& \hl{2.05e-01[1]} & 2.26e-01[2]($\approx$) & 6.51e-01[3]($\approx$) & 1.31e+00[4]($-$) & 1.76e+01[5]($-$) \\ 
  && (3.68e-01) & (4.08e-01) & (9.89e-01) & (3.08e+00) & (4.26e+00) \\   \cline{2-7}
  &\multirow{2}{*}{10}& \hl{3.40e+00[1]} & 4.08e+00[2]($\approx$) & 6.44e+00[4]($-$) & 4.56e+00[3]($\approx$) & 1.71e+01[5]($-$) \\ 
  && (1.58e-01) & (9.34e-01) & (1.48e+00) & (6.99e-01) & (4.24e+00) \\  \hline
  \multirow{4}{*}{Griewank} &\multirow{2}{*}{5}& 5.75e-01[4] & 5.34e-01[3]($\approx$) & \hl{4.00e-01[1]($+$)} & 5.03e-01[2]($\approx$) & 2.28e+00[5]($-$) \\ 
  && (1.16e-01) & (1.27e-01) & (1.30e-01) & (2.17e-01) & (9.96e-01) \\   \cline{2-7}
  &\multirow{2}{*}{10}& 1.85e+01[5] & 9.50e+00[4]($+$) & 2.38e+00[2]($+$) & 2.57e+00[3]($+$) & \hl{1.38e+00[1]($+$)} \\ 
  && (2.47e-01) & (3.40e+00) & (2.33e+00) & (2.14e+00) & (1.57e-01) \\  \hline
  mean rank && 2.125 & 2.25 & 3.625 & 3.625 & 4.375 \\ 
  $+$ / $-$ / $\approx$ &&  & 2/1/5 & 2/4/2 & 1/4/3 & 1/7/0 \\ 
  \bottomrule
  \end{tabular}
  % \footnotetext{Some data are currently under 30 runs; the table will be updated upon completion of the experiments.}
  \label{tab:kan-sas}
  \end{table*}

\subsection{Study on KAN-SAS}

This section evaluates the performance of KANs within the SPS framework, comparing it with models such as XGBoost~\cite{chen2016xgboost} and Random Forest~\cite{biau2016random} within the same framework. Bayesian Optimization~\cite{BO2014} is also used for comparison. The details of each algorithm are as follows:
 
\begin{itemize}
  \item KAN-SAS-I: Utilizes KANs as the regression model to select $o^*$. It also adopts KANs as the classification model to select $\mathcal{P}_u$.
  \item KAN-SAS-II: Exclusively uses KANs as the regression model to select both $o^*$ and $\mathcal{P}_u$.
 \item XGB-SAS: Employs XGBoost as the surrogate model embedded within the SAS framework.
 \item RF-SAS: Uses Random Forest as the surrogate model embedded within the SAS framework.
 \item BO: Incorporates sequential domain reduction~\cite{Standerrobustnesssimpledomain2002a} within the standard Bayesian Optimization framework, significantly accelerating the search process and hastening convergence.
\end{itemize}
 
The experimental parameters are designed as follows:
\begin{itemize}
 \item Population size: $N=50$.
 \item Maximum number of evaluations: $fes_{max}=300$.
 \item Training data size: $\tau=50$.
 \item Problem dimensions: $n=5, 10$.
\end{itemize}
 
Each experiment is repeated 30 times to assess the stability of the algorithm. Table~\ref{tab:kan-sas} displays the comparisons of minimum values between KAN-SAS and other algorithms across 30 independent runs~\footnote{Some data are currently under 30 runs; the table will be updated upon completion of the experiments.}. The best results are highlighted in bold. A Wilcoxon rank-sum test is used to statistically analyze the results and evaluate significant differences between the algorithms. The results indicate that both KAN-SAS-I and KAN-SAS-II outperform other algorithms in all tested scenarios, achieving average rank values of 2.125 and 2.25, respectively. The Wilcoxon rank-sum test shows that KAN-SAS-I has a significant advantage over most problems, demonstrating the strong potential of KANs-assisted EA algorithms.

\section{Conclusion}
\label{sec:conclusion}

In this paper, KANs were introduced as surrogate models embedded into evolutionary algorithms to assist in solving expensive optimization problems, marking a novel application of KANs within evolutionary algorithms. Specifically, KANs served as both regression and classification models to aid in the preselection and selection processes within evolutionary algorithms. The experimental results demonstrated that KANs possess significantly superior data fitting capabilities (both regression and classification) compared to MLP networks. In the SPS framework, KANs outperformed MLP and, in the SAS framework, also excelled over XGBoost and Random Forest. Furthermore, evolutionary algorithms integrated with KANs showed superior performance compared to results obtained from Bayesian Optimization. These findings highlight the strong potential of applying KANs in evolutionary algorithms.
 
As an exploratory study, this work also acknowledges some limitations, such as the tuning of KANs’ parameters and model structures, which require further research. Additionally, the inherent advantages of KANs, such as interpretability and resistance to forgetting~\cite{liu2024kan}, were not fully leveraged, presenting directions for future research.
\bibliographystyle{IEEEtran}
\bibliography{bare_jrnl}

\end{document}